# Trees from Marginals: Autoregressive drafting with factorized priors

**Yuma Oda, Ryan Mathieu, Roman Knyazhitskiy** and **Artur Chakhvadze**[1]

mirai labs

## Abstract

Speculative decoding greatly increases the interactivity of autoregressive language models by trading off computation for extra tokens generated in a single forward pass. Factorized draft models are especially efficient because they predict future-token marginals in parallel, but their independence assumption causes acceptance rates to degrade sharply at later draft positions. We analyze this limitation and introduce Weaver, a lightweight autoregressive adapter that constructs proposal trees from the top-$K$ prediction of a factorized drafter. Weaver restores conditional dependencies between proposed tokens while avoiding the cost of a full-vocabulary projection. To support fast verification for models with Gated Delta Net layers, we derive a rollback-free tree-verification algorithm and implement optimized CUDA kernels in SGLang. By combining these model and systems contributions we achieve a 4.08× speedup over autoregressive decoding, and outperform a highly optimized DFlash baseline by 16.3%.

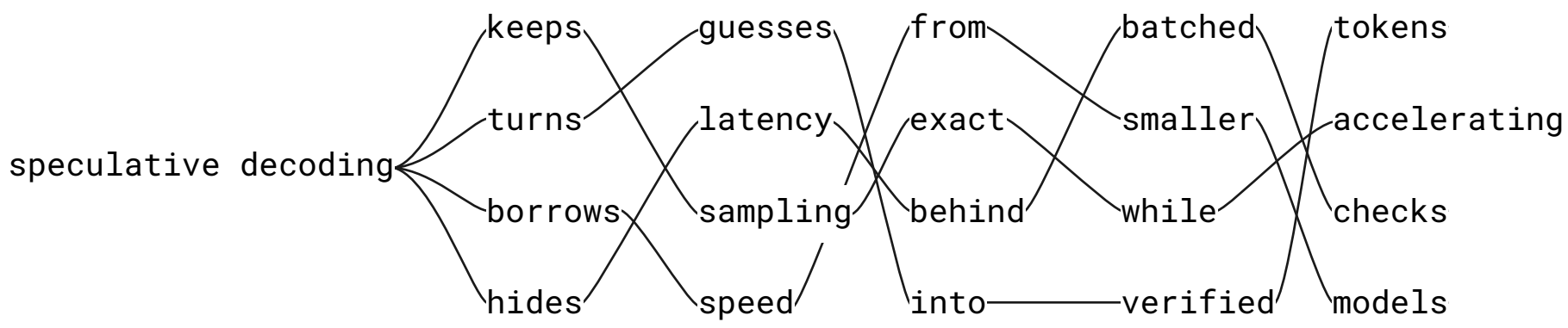

[1]Correspondence to artur@getmirai.co

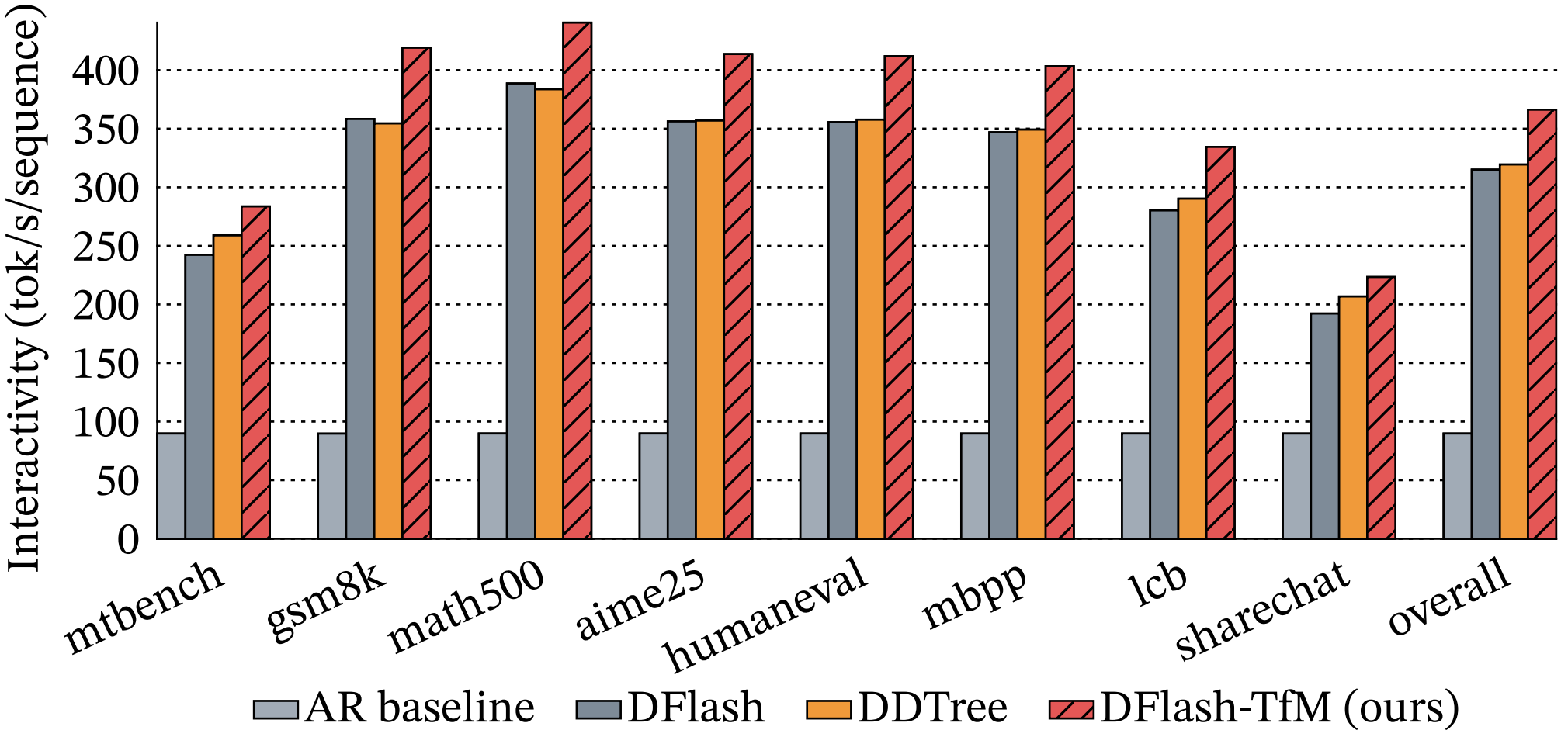

Figure 1: Comparison of decoding speed for Qwen3.6-27B in SGLang in bfloat16 precision on different datasets with default sampling parameters (temperature 1.0, reasoning on). Our method outperforms an optimized DFlash baseline by 16.3% on average.

# 1 Introduction

## 1.1 Motivation

The dominant neural architecture powering large language models is an autoregressive transformer[1]. It allows for parallel likelihood evaluation and training, but generation remains sequential, since the generated sequence has to be sampled autoregressively one token per step.

For a given hardware setup, the time spent on each generation step is a function of communication and computation cost. Communication involves moving the weights, activations and KV cache between DRAM and SRAM[1], while computation consists mostly of matrix multiplications and dot-product attention operations. A reasonably good inference implementation can overlap computation and communication to a high degree, so the overall time spent on a single decoding step can be well approximated as

$$\text{latency} = \max\left(\frac{\text{computation cost}}{\text{arithmetic performance}}, \frac{\text{communication cost}}{\text{memory bandwidth}}\right)$$

[1]For the sake of simplicity we only consider single-GPU deployments in this exposition.

When generating a single sequence, this latency is dominated by the memory traffic due to streaming weights from DRAM, which makes generation *memory-bound*. This is undesirable because of compute underutilization and poor energy efficiency per token, as DRAM accesses consume orders of magnitude more energy than arithmetic operations[2].

In order to increase the ratio of computations to memory traffic, server LLM inference implementations make use of *batching* – decoding multiple sequences in parallel. Batching increases the cost of computation while keeping the weight traffic constant. Unfortunately, it is not a free lunch, as larger batch sizes result in increased activation and KV cache traffic, and the latter quickly becomes the bottleneck for long sequences.

Additionally, increased batch size reduces *interactivity*[1], which is critical for user-facing applications and asynchronous reinforcement learning algorithms, where the policy gradient bias depends on the speed of rollout completion[3], [4].

*Speculative decoding*[5] is an orthogonal approach for increasing the arithmetic intensity of generation while simultaneously increasing interactivity and maintaining constant KV cache traffic. The idea is to use a fast auxiliary *draft model* to generate a sequence of proposal tokens, and then compute their likelihood using the main model (called *verifier* hereafter) in parallel in a single step. A procedure similar to rejection sampling[2] is then used to accept or reject each proposed token individually based on the likelihood ratio of drafter and verifier. This way, multiple tokens can be generated in a single step, effectively parallelizing generation along the sequence dimension.

Local LLM setups require high interactivity and energy efficiency, but can rarely benefit from batching, since most local AI use cases involve processing only a single query at a time. Therefore, speculative decoding methods which scale well to large speculative budgets are especially important for local LLM inference.

Recently, factorized models inspired by discrete diffusion have been used to set new records in speculative decoding efficiency at small speculative budgets[6], [7], [8], [9], as factorization enables fast parallel generation of proposals even with very large drafters. Unfortunately, marginal token distributions diverge from true autoregressive conditionals as draft length increases, which puts hard limits on acceptance rates of factorized drafters, and makes them scale poorly with increased draft lengths, as we show in Section 3.5.

[1] We define interactivity as the number of tokens decoded per second per every sequence in batch.

[2] While being structurally similar, speculative sampling is **not** the same as rejection sampling. The two methods are often conflated, even by people working on inference engines, which leads to significant confusion. See Appendix A.2. in Leviathan et al.[5] for more details.

### 1.2 Our contributions

In this work we introduce **DFlash-TfM**, a hybrid draft model architecture which achieves excellent scalability properties by combining the benefits of factorized and autoregressive drafters.

We use the top-$K$ marginal predictions from a pre-trained factorized DFlash[9] model as a prior for a small autoregressive drafter which we call *Weaver*. We constrain the candidate vocabulary at each speculative position to the top-$K$ marginal tokens to avoid the expensive multiplications by the output projection matrix, keeping the autoregressive decoding overhead to the minimum.

By using **Weaver**[1] to construct trees of proposals and verifying them using the traversal verification algorithm[10], we increase the mean acceptance length by 73.6% relative to the chain DFlash baseline, and by 22.3% relative to DDTree[11] with the same tree size.

We analyze the behavior of a theoretically optimal factorized draft model and show that the acceptance rates of our hybrid model for long draft sequences are higher than what can be achieved by any drafter trained to predict the marginal per-token distributions.

In order to use our method with Qwen3.6 family of models, we derive a novel algorithm for efficient tree verification with Gated Delta Net layers, and implement efficient CUDA kernels for the SGLang framework[12].

Together, all of these contributions enable 16.3% improvement in tokens/s/sequence over an optimized DFlash baseline on a single B200 in SGLang, using Qwen3.6-27B model in bf16 precision.

[1]Weaver weights are available on **Hugging Face**. The reference implementation is available in the **SGLang fork on GitHub**.

# 2 Background and related work

## 2.1 Drafter architectures

Draft models proposed in the literature can be organized along two axes.

The first concerns where the drafter's parameters live: a *standalone drafter* is a separate model with its own parameters, coupled to the target only through a shared vocabulary[5], [13], whereas an *adapter* introduces no separate model and instead attaches to and reuses the target's weights, hidden states, or output projection matrix head[14], [15].

The second axis concerns how a multi-token proposal is produced. An autoregressive drafter emits draft tokens sequentially, each conditioned on those before it, so the proposal follows the chain rule and diverges from the target only through drafter error. A factorized drafter instead predicts several future positions in a single forward pass as conditionally independent marginals given the prefix, trading exactness for parallelism.

Alternatively, there exists a family of training-free methods that dispenses with a parametric drafter altogether and proposes tokens by string matching or self-drafting[16], [17], [18].

### 2.1.1 Autoregressive drafters

The original speculative-decoding formulation used a small standalone draft model. *Self-speculative* methods remove the separate model by generating drafts using a pruned version of the verifier: *Draft & Verify*[19] drafts by skipping a subset of the target's intermediate layers, with no auxiliary parameters and no extra training, while *LayerSkip*[20] drafts via early exit at an intermediate layer, enabled by a layer-dropout and early-exit training recipe; both share KV cache and activations between drafting and verification.

A second line of work realizes the drafter as a lightweight adapter on top of the features extracted by the verifier model. *Hydra*[21], *EAGLE*[15], *EAGLE-2*[22] and *EAGLE-3*[23] all train lightweight adapters which reuse the residual stream activations of the verifier model and share its output projection matrix. To reduce the memory traffic due to multiplication by the final vocabulary projection, Gemma4-MTP[24] groups token embeddings into clusters, and constraints the output distribution only to top-$K$ clusters whose centroids are closest to the final embedding.

### 2.1.2 Factorized drafters

Factorized drafters predict $T$ future positions in a single forward pass, drawing each from a marginal that is conditionally independent of the other drafted positions given the prefix.

*PARD*[6] and *PARD-2*[8] adapt a low-cost autoregressive draft model into a parallel one with mask-token placeholders.

*Medusa*[14] adds lightweight decoding heads on top of the last layer features of the verifier, with all heads predicting future tokens in parallel from the same hidden state. *DFlash*[9] uses a large transformer-based adapter that decodes an entire block in one pass using non-causal in-block attention over mask tokens, with verifier hidden states injected into the KV cache of the draft model.

*I-DLM*[7] fine-tunes the verifier to predict marginal distributions over future tokens in parallel when given special future-mask tokens as inputs, and uses the same model for factorized drafting and autoregressive verification.

Although these methods inherit the *diffusion* lineage, at draft time they operate as single-step parallel masked-marginal predictors rather than multi-step denoising, since multi-step denoising would make the draft likelihood intractable and speculative verification impossible.

Factorization is what lets these drafters set records at small speculative budgets, but the same independence assumption imposes a structural ceiling. Because position $n + k$ is drawn from a marginal that ignores the realized values at $n \ldots n + k - 1$, the draft distribution diverges from the target's chain-rule conditionals, and this divergence grows with draft length: per-position acceptance decays with depth, capping the useful block size and the expected accepted length.

## 2.2 Verification methods

A naive approach to speculative decoding would be to sample a sequence of draft tokens $\tilde{x}_i$ from the draft model

$$(\tilde{x}_1, \ldots, \tilde{x}_T) \sim p_{\text{drafter}}(\ldots \mid c) \tag{1}$$

and then use them as conditionals to generate parallel independent samples from the verifier

$$x \sim p_{\text{verifier}}(\cdot \mid c, \tilde{x}_{<i}) \tag{2}$$

and emit the first $k$ matching tokens ($x = \tilde{x}_i$), where $p_{\text{drafter}}$ and $p_{\text{verifier}}$ are respectively drafter and verifier output distributions, when they are provided with the context $c$.

We call this algorithm *naive verification*. Although it is optimal for greedy decoding, it has a fairly poor acceptance rate in general, since it doesn't guarantee acceptance even when the drafting and verification are performed by the same model.

#### 2.2.1 Speculative sampling

*Speculative sampling* is the original sampling scheme proposed by Leviathan et al.[5] and independently by Chen et al.[13] under the same name. Instead of sampling tokens from the verifier separately, it attempts to accept draft tokens one by one by performing a likelihood ratio test.

$$p_{\text{accept}} = \min\left(\frac{p_{\text{verifier}}}{p_{\text{drafter}}}, 1\right) \tag{3}$$

Acceptance probability for speculative sampling can be also expressed in terms of the total variation distance between the drafter and verifier distributions:

$$p_{\text{accept}} = 1 - \mathrm{D}_{\mathrm{TV}}(p_{\text{drafter}}, p_{\text{verifier}}). \tag{4}$$

If the drafter and verifier perfectly agree, the proposals are always accepted, and the maximum coupling theorem for Markov chains[25] implies that this, in fact, is the best achievable acceptance rate for single-step drafts.

#### 2.2.2 Block verification

*BlockVerify*[26] improves the verification procedure for multi-token drafts by searching for an optimal transport coupling between joint distributions over sequences of tokens.

Instead of verifying tokens one by one, it attempts to jointly accept a draft prefix of length $k - i + 1$ for every $i$ from 1 to $n$.

#### 2.2.3 Tree verification

Since the probability of accepting a token drops quickly with its position in the draft sequence, increasing the length of the draft provides diminishing returns.

To make better use of larger draft budgets, tree-based speculative decoding methods[14], [27] construct a prefix tree from multiple proposals. A simple modification to the causal attention mask (usually referred to as tree attention) allows to efficiently evaluate the verifier likelihood of every token in the tree in parallel.

The original *SpecInfer*[27] implementation constructs the tree of proposals by sampling branches independently with replacement and uses speculative sampling as a verification method. *Sequoia*[28] improves upon it by drawing children jointly without replacement when extending a branch, and applying the corresponding adjustment to the residual correction in the speculative sampling procedure.

*Traversal verification*[10] extends the BlockVerify block-level verification idea to trees of tokens, and *SpecTr*[29] uses optimal transport framework to optimize multi-candidate per-token acceptance; they note that lossless verification is computationally unfeasible, and resort to an approximation with strict bounds on optimality. Khisti et al.[30] showed that any OT-based tree verification method can be represented as a two-step algorithm where the first step selects a branch and the second step performs speculative sampling.

#### 2.2.4 Tree shape optimization

The structure of the draft tree has a major impact on the performance of speculative decoding. Early methods such as SpecInfer[27] and EAGLE[15] used heuristic-based static tree structures. Sequoia[28] proposed an optimization method to statically select an optimal tree shape for a given pair of drafter and verifier. *DySpec*[31] proposed a dynamic programming algorithm to optimize the tree shape for every generation step in runtime.

#### 2.2.5 Tree verification for recurrent models

Tree verification as described above presupposes attention: scoring the whole draft tree in one pass reduces to an ancestor-restricted attention mask[27], a reduction that recurrent layers do not admit. Speculative decoding for recurrent targets was accordingly first developed with linear chain drafts[32], [33], which sidestep branching at the cost of the diminishing returns of long chains. *STree*[34] brought tree verification to state-space models by exploiting the accumulated state-transition matrices of the draft tree: for a diagonal, Mamba-style transition the state at a node is a product of the gates along its path, so the entire tree is scored by an ancestor-masked scan with no triangular solve. This reduction is specific to diagonal recurrences – a gated delta net[35] advances its state by the non-commuting matrix $I - \beta_t k_t k_t^\top$, which admits no cumulative-product form, leaving efficient single-pass tree verification for delta-rule targets an open problem.

# 3 Method

## 3.1 Trees from marginals

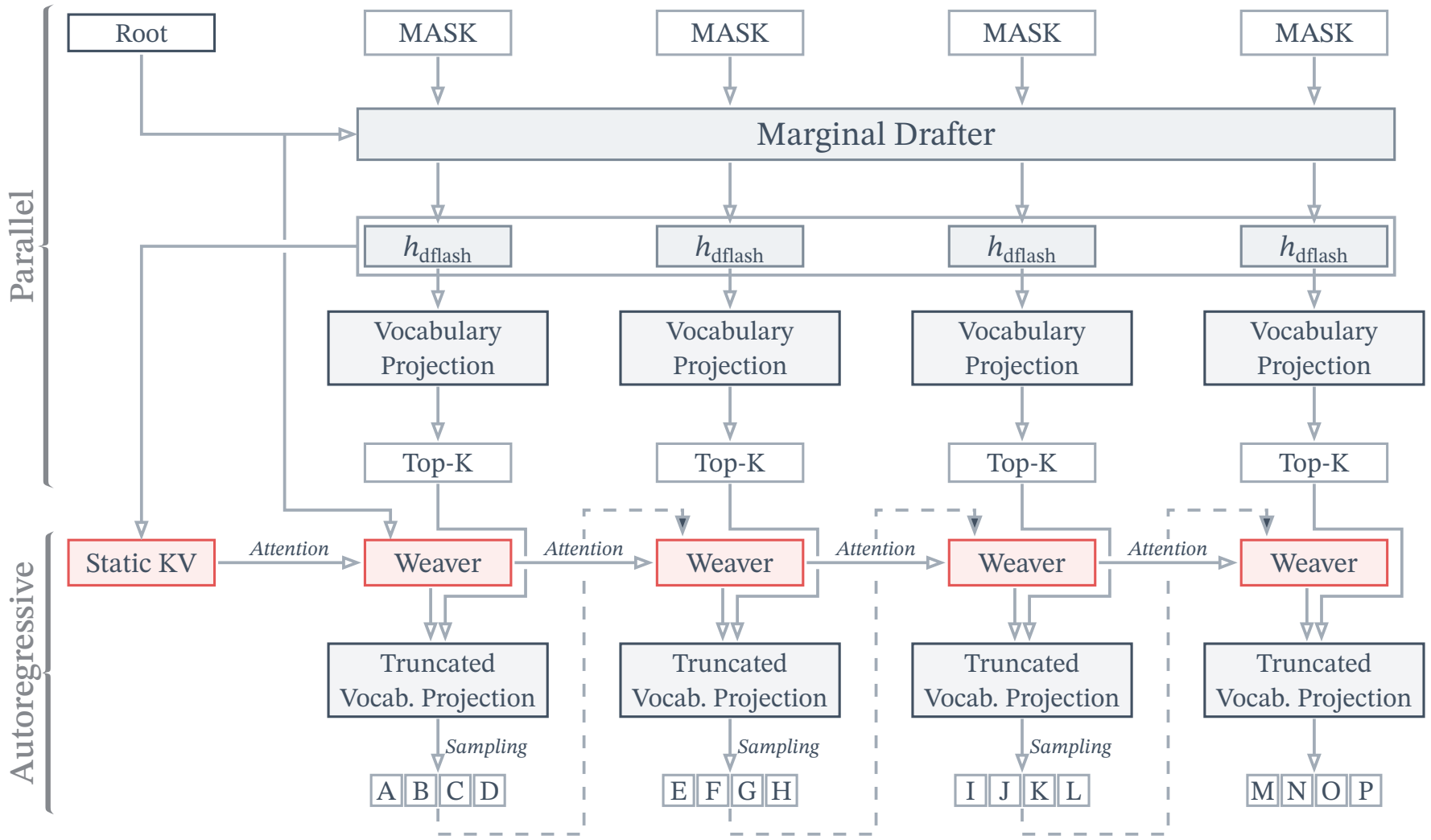

Figure 2: Overview of the DFlash-TfM drafting procedure.

A factorized drafter emits marginal distributions for several positions in a single forward pass. Because the factorized drafter ignores conditional dependencies between positions, the acceptance rate falls as the draft length grows (Figure 4). An autoregressive drafter captures these dependencies correctly, but each proposal incurs the cost of a full-vocabulary projection. Therefore, gains in quality translate directly into drafting overhead. Moreover, for targets built primarily on Gated Delta Net layers, such as Qwen3.6, the sequence ancestor mask does not hold because of the non-commutative state transition. This limitation leaves fast tree verification an open problem.

Our method, **DFlash-TfM** (shorthand for Trees from Marginals)[1], addresses

[1] The proposed method is not constrained to DFlash; the model was chosen as the current state-of-the-art marginal drafter. In practice one can use our method on top of any other par-

these constraints. We extract the top-$K$ marginal predictions produced in a single forward pass by the factorized drafter DFlash[9]. We then employ a light-weight autoregressive draft model, **Weaver**, which operates exclusively over that candidate vocabulary (which is a tiny subset of the full vocabulary) to restore the dependencies conditioned on the previously selected draft tokens. Weaver shares the embedding matrices and the output projection matrix of the verifier, but the network never projects to the full vocabulary. This allows the model to avoid the memory-bandwidth cost of a standard autoregressive drafter.

The proposal tree constructed by Weaver is verified in one pass. For targets that include GDN layers, we introduce a dedicated kernel that verifies an entire tree without rolling the recurrent state back, detailed in a Section 3.4.

## 3.2 Weaver architecture

Weaver is a lightweight autoregressive transformer that learns to predict the residual between the marginal distributions predicted by a factorized drafter and the verifier output distribution.

Operating similarly to EAGLE [23] , it takes the intermediate activations from the verifier, the *future state lookaheads* from the parallel drafter, and the generated tokens to predict the autoregressive residual.

At the start of a new speculative decoding round, DFlash takes in the last token verifier hidden state $h_{\text{verifier}}$ to produce a sequence of future state lookaheads $h^1_{\text{dflash}}, \ldots, h^L_{\text{dflash}}$. Multiplying these states by the verifier vocabulary projection matrix $W_{\text{vocab}}$ yields the baseline marginal logits: $\ell^i_{\text{dflash}} = W_{\text{vocab}} h^i_{\text{dflash}}$. In the original DFlash the tokens sampled using these logits *are* the draft tokens. We, however, use these logits as the prior for the autoregressive tree construction.

Weaver takes in both the hidden state of the verifier $h_{\text{verifier}}$ and the hidden states $h^{1..L}_{\text{dflash}}$ of the parallel drafter, starting by compressing them into conditioning tokens:

$$\begin{aligned} u^0 &= W_{\text{c}} \, \text{RMSNorm}\left(h_{\text{verifier}}\right) \\ u^i &= W_{\text{c}} \, \text{RMSNorm}\left(h^i_{\text{dflash}}\right) + \text{p}^i \end{aligned} \tag{5}$$

where $W_c$ is the conditioning weight matrix trained from scratch, and $p^i$ is allel drafter, e.g. I-DLM[7].

positional encoding. The sequence of vectors $u$ are the continuous tokens, the *prompt*, from which the static KV cache is constructed.

For instance, expanding a tree path with tokens $t^{1..d}$, the Weaver input would be the concatenation of the prompt tokens $u^{0..L}$ and the tokens along the draft tree path $t^{1..d}$.

The new logits are generated via a pass through a small transformer WeaverStep:

$$\ell_{\text{draft}}(u^{0..L}, t^{1..d}) = \text{WeaverStep}(u^{0..L}, t^{1..d}) \tag{6}$$

To convert the residual stream into logits we need to multiply them by the vocabulary projection matrix. Most of the memory bandwidth pressure of Weaver execution comes from reading this matrix. We reduce the communication necessary by only reading $K$ rows of this matrix (usually 512), selected as Top-$K$ DFlash tokens, adding Weaver residual logits to the DFlash output logits, and then normalizing over the candidate set.

### 3.3 Tree construction

DySpec[31] demonstrated that constructing maximum-drafter-probability[1] trees requires expanding the highest-probability node in a best-first structure. An issue with the DySpec approach is that enforcing strict sequential tree construction is slow under highly parallel execution.

We alter the DySpec algorithm to allow for better memory bandwidth utilization. Instead of selecting the single best node from the heap of draft-tree candidate nodes and executing a single frontier expansion, we extract the top $w$ nodes from the heap and perform $w$ corresponding expansions concurrently.

Because batched execution allows to balance bandwidth and compute pressure, evaluating a Weaver batch of size $w$ incurs a latency nearly identical to evaluating an unbatched sequential step. Assuming a total token budget $B$, parallel node extraction constructs the complete draft tree in $\lceil B/w \rceil$ sequential operations.

This modification presents a trade-off. The multi-node frontier expansion includes sub-optimal draft-probability nodes, however it improves drafting throughput significantly by eliminating the inefficiency of sequential Weaver

[1]Maximum-draft-probability is a useful proxy for estimating the acceptance probability, but it is imperfect. For a detailed discussion see DySpec[31].

runs. We find that optimal $w$ lies between $2 - 8$.

## 3.4 Optimized tree verification kernel for gated delta net

Verifying proposal trees with pure attention-based models is relatively straightforward and involves a simple modification of the attention mask[15], [27]. More recent target models utilize state-space layers, such as gated delta networks (GDN)[35], which keep their context in a running recurrent state. The masking approach does not transfer to such layers directly. Prior methods[34] perform sequential scans along tree branches, which scale poorly with the size of the tree.

Our verification procedure instead uses the dual chunk form of the linear recurrence. We avoid speculatively updating the state during verification, and delay the state commit until the next decoding iteration when the accepted branch is known. On commit we replay the selected path using a short recurrence. The read-only verification step reduces to a masked triangular solve, which we fuse into a single GPU kernel.

### 3.4.1 Chunked Gated Delta Rule for tree-structured drafts

A GDN layer processes a sequence of tokens by maintaining a state matrix $S_t \in \mathbb{R}^{d_k \times d_v}$.

At each step $t$, it takes a pair of key and value vectors $k_t \in \mathbb{R}^{d_k}$, $v_t \in \mathbb{R}^{d_v}$, as well as a decay factor $\alpha_t \in (0, 1]$ and a write strength factor $\beta_t \in (0, 1]$, and performs a state update:

$$S_t = \alpha_t \big(I - \beta_t k_t k_t^\top\big) S_{t-1} + \beta_t k_t v_t^\top \tag{7}$$

It then uses a query vector $q_t \in \mathbb{R}^{d_k}$ to extract the output $o_t$ from the updated state:

$$o_t = \frac{q_t^\top S_t}{\sqrt{d_k}} \tag{8}$$

The chunk form of the delta rule allows to compute the result of $n$ consecutive updates in a single pass via a pair of triangular solves[35], [36]. We derive a simple extension of the chunked form which operates on trees of tokens.

Let $\prec$ be an ancestor relation in a draft tree, then token $i$ can only receive context from token $j$ if $j \prec i$. We sort tokens in a draft tree in the topological order, such that $i \preceq j \Rightarrow i \leq j$

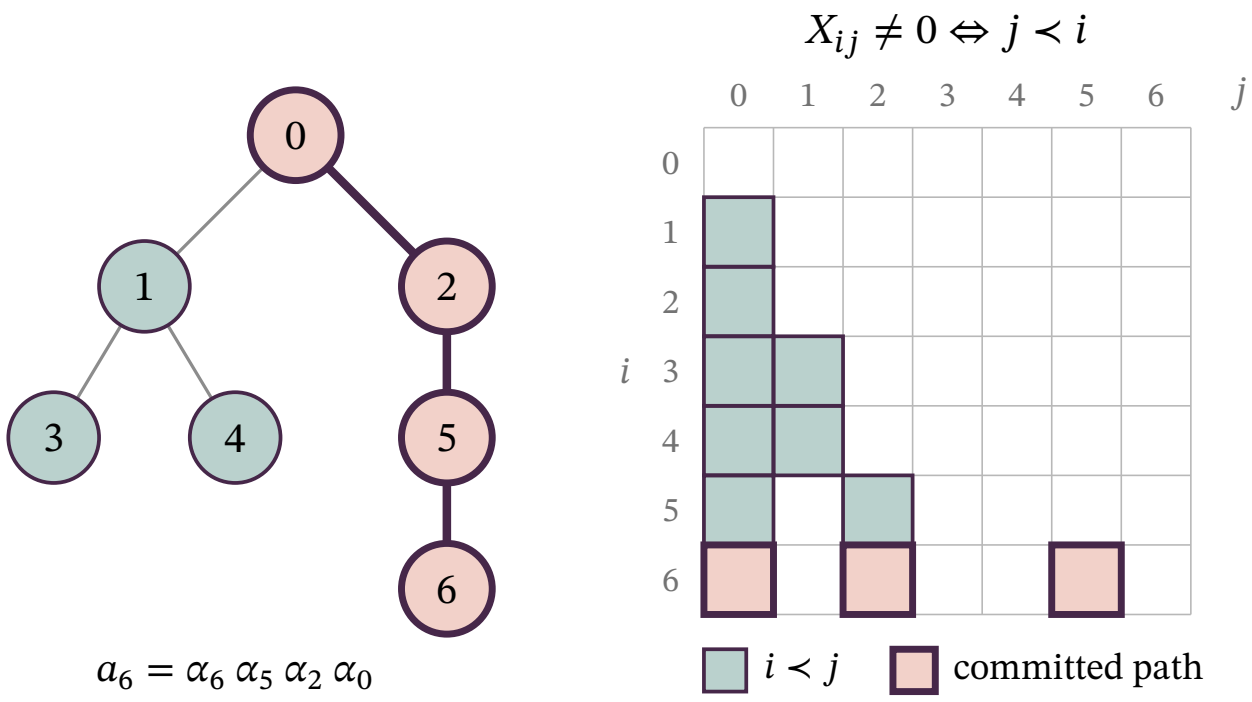

Figure 3: Tree verification modifies the GDN chunk algebra to use a partial order: a token at index $i$ interacts only with its ancestors $j \prec i$.

Let matrices $K \in \mathbb{R}^{n \times d_k}$ and $V \in \mathbb{R}^{n \times d_v}$ be row-stacked key and value vectors, and let $a_t = \prod_{i \preceq t} \alpha_i$ be a sequence of cumulative products of decay factors along tree branches. The cumulative key-key and query-key interactions can be represented by lower-triangular matrices $X \in \mathbb{R}^{n \times n}$ and $Y \in \mathbb{R}^{n \times n}$ respectively:

$$X_{ij} = \mathbb{I}[j \prec i]\, \frac{a_i}{a_j} \beta_i k_i^\top k_j, \quad Y_{ij} = \mathbb{I}[j \preceq i]\, \frac{a_i}{a_j} q_i^\top k_j \tag{9}$$

Following Yang et al.[35], we define auxiliary matrices $W$ and $U$ as solutions to linear systems

$$(I + X)U = \beta V, \quad (I + X)W = \beta a K \tag{10}$$

where $\beta = \operatorname{diag}(\beta_t)$ and $a = \operatorname{diag}(a_i)$.

Then the outputs of the GDN layer in the form of a matrix $O \in \mathbb{R}^{n \times d_v}$ can be computed as

$$O = \frac{1}{\sqrt{d_k}} (aQS_0 + Y(U - WS_0)) \tag{11}$$

The same objects give the state at the end of the block. For the accepted leaf $n$,

$$S_n = a_n S_0 + K^\top (a_n a^{-1})(U - WS_0), \tag{12}$$

where $a_n a^{-1}$ scales row $j$ by $a_n / a_j$: each token contributes its corrected write,

decayed by the remaining decay to the end of the path. We instead replay the accepted path with Equation 7 at commit time, since the fused kernel folds $WS_0$ into the solve and never materializes $W$.

### 3.4.2 Batched fused-kernel tree implementation

The proposal tree is shared by every GDN layer, so its verification metadata is computed once per decode step. From the parent pointers we compute, for each node, the cumulative decay $a_i$ (accumulated in log space) and the set of its proper ancestors; the ancestor sets are stored as bitmasks, one bit per candidate ancestor, packed into 64-bit words. The kernel inputs are dense tensors of shape $[N, T, H, d]$, with $N$ the batch, $T$ the padded tree size, and $H$ the number of key heads. These shapes do not depend on the tree's contents, so the whole pass is captured once and replayed inside a CUDA graph. The kernel computes [scalars], [write strengths $\beta$ and cumulative decays $a$, for all heads at once], [3.4], [4.9], builds the interaction matrix from the Equation 9 per key head, and runs the forward substitution tiled into sub-blocks of $B_c = 32$ nodes. Each diagonal block $(I + X_{bb})$ is inverted in registers by repeated squaring, and a cascade over sub-blocks handles the off-diagonal coupling,

$$U_b = (I + X_{bb})^{-1}\left((\beta \odot V)_b - \sum_{c<b} X_{bc}\, U_c\right), \tag{13}$$

which is $\lceil T/B_c \rceil$ dense block steps rather than $T$ scalar steps. The $-w_j^\top S_0$ term of Equation 11 is folded into the right-hand side of the solve, so $W$ is never formed and the substitution returns $u_j - w_j^\top S_0$ directly; the output stage then reads $S_0$ once, decayed, and applies the ancestor-masked attention against the solve result.

We lay the state out as $[H_v, d_v, d_k]$ to match the decode and prefill GDN kernels ($H_v$ is the number of value heads). All matrix products use `tl.dot` with tf32 inputs and fp32 accumulation, except for the sensitive log-decay accumulations, which use tf32x3 ($\sim 5 \times 10^{-7}$ relative); the precision policy keeps the verify within $10^{-4}$ of the double-precision reference. Once Traversal verification picks the accepted leaf, a single commit pass replays Equation 7 along that path and writes $S_0$, which is the only state write in the entire decode step.

| stage | computes | $T = 64$ | $T = 128$ |
|---|---|---|---|
| scalars | write strengths $\beta$ and path sums $G$, for all heads at once | 3.4 | 4.9 |
| Gram | the interaction matrix $X$ per key head, and the inverse of each diagonal sub-block | 12.2 | 16.7 |
| solve | the forward substitution (Equation 13) | 6.5 | 13.8 |
| output | the block output Equation 11 | 5.7 | 9.3 |
| total | | **27.8** | **44.7** |

Table 1: Stages of the fused verify kernel and their self-time (µs, B200, batch size 1). The Gram and the solve dominate, and the solve is the one stage whose cost grows super-linearly with $T$ (a factor of 2.1 from $T = 64$ to 128). The self-times sum to the compute portion; the rest of the Figure 5 wall-clock is kernel launch and inter-stage copies, fixed under CUDA-graph replay.

### 3.5 Empirical performance bounds of factorized drafting

The expected acceptance rate can be upper bounded by using the total variation distance between proposal and target distributions. For the simplicity of analysis we consider only speculative sampling. During chain speculative sampling the token-wise acceptance probability is exactly $1 - \mathrm{D_{TV}}(p_{\text{draft}}, p_{\text{verifier}})$.

For a future position $L + t$ where $L$ is the index of the last verified token, an oracle marginal drafter would use the true future-token marginal[1]

$$p(x_{L+t} \mid x_{\le L}) = \mathbb{E}\left[p_{\text{verifier}}(x_{L+t} \mid x_{<L+t}) \mid x_{\le L}\right] \tag{14}$$

We estimate those marginal probabilities to get an upper bound for the acceptance rate of an oracle marginal drafter. For each context sequence of length $L$, we sample $M$ random continuations $x^m$ from the target model. We then run the target model on the produced token sequence and compute the output token distribution $p(x_{L+t} \mid x^m_{<L+t})$ at each position.

We then estimate the position-wise marginal probabilities as

[1]In principle one could train a factorized drafter to minimize the expected TV directly, but standard training objectives match marginals.

$$\hat{p}(x_{L+t} \mid x_{<L}) = \frac{1}{M} \sum_{m=1}^{M} p\left(x_{L+t} \mid x_{<L}, x_{<L+t}^{m}\right). \tag{15}$$

The acceptance rate at a future position $t$ can be then estimated as

$$p_{\text{accept}}(t) = 1 - \mathbb{E}\left[\mathrm{D}_{\text{TV}}\left(\hat{p}(\cdot \mid x_{<L}), p\left(\cdot \mid x_{\leq L}, x_{<L+t}^{m}\right) \mid x_{\leq L}\right]\right. \tag{16}$$

where expectation is taken over the context sequences and continuations in the dataset.

We estimate the values of these bounds from the MTBench[37] dataset, and observe that at later positions our proposed architecture, DFlash-TfM, achieves acceptance probabilities higher than the that of any marginal or argmax-marginal drafter.

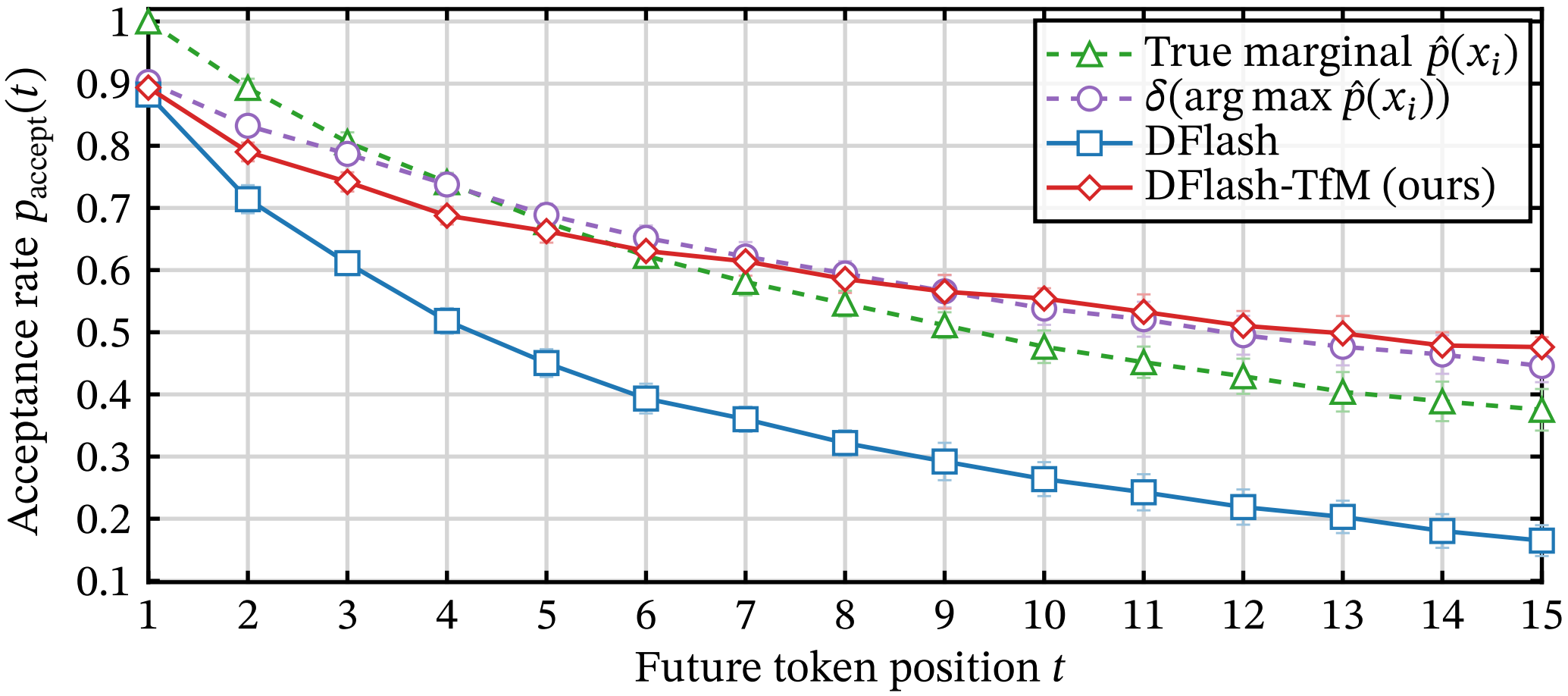

Figure 4: Acceptance rates under speculative sampling computed for different draft distributions. Error bars show 95% confidence intervals

Standard training objectives for diffusion drafters naturally target matching of the per-token marginal probabilities. However, doing so does not automatically maximize the expected acceptance rate, and in some settings optimizing for a different target could be better.

For example, using a Dirac delta distribution with the peak at the top-1 token of the marginal distribution (equivalent to using naive verification and argmax sampling during drafting) empirically outperforms the full marginal distribution on later positions even for non-greedy decoding, which results in better acceptance probabilities for longer chains. This explains the results in Table 3, where naive verification with argmax draft sampling outperforms speculative sampling for DFlash baseline.

This result also suggests that for optimal performance of purely factorized drafters, it is best to adaptively anneal the distribution by using different temperatures at different token positions. We leave the exploration of this idea for future work.

# 4 Implementation details

## 4.1 Training implementation

We train Weaver on top of the publicly released Qwen3.6-27B-DFlash[1] checkpoint, producing an autoregressive drafter as the only artifact.

### 4.1.1 Rollout generation

For training, Weaver needs only the verifier logits for the loss and the DFlash lookaheads for its input, but storing these across a large rollout has a prohibitively large on-disk footprint, so we first run an offline rollout – storing only the token ids – and then recompute both logits and lookaheads during training. We generate verifier responses at temperature 0.0, reasoning on, with responses of up to 8192 tokens. In a later run, each response is prefilled into the verifier to recover its logits together with the context features DFlash injects into its KV cache; running DFlash over these features then yields the output future state lookaheads Weaver consumes.

### 4.1.2 Data augmentation

Following DFlash[9], we resample anchor positions from each response sequence, where an anchor is the position of the verified bonus token the drafter conditions on at inference. We adapt this augmentation scheme to the autoregressive setting and sample 128 anchor positions per sequence.

### 4.1.3 Training objective

We train Weaver with the LK loss[38] between the verifier distribution $p$ and the proposal distribution $q$,

[1] https://huggingface.co/z-lab/Qwen3.6-27B-DFlash

$$L_{\mathrm{LK}}^{\lambda}(p, q) = \lambda \cdot \mathrm{KL}(p \parallel q) + (1 - \lambda) \cdot \mathrm{TV}(p, q), \quad (17)$$

whose mixing weight $\lambda = \exp(-\eta \cdot \mathrm{stop_gradient}(1 - \mathrm{TV}(p, q)))$ is computed per position. While the student is far from the teacher, the loss acts as a forward KL; as it converges, $\lambda$ decays and the weight shifts onto the TV term, which is equal to the $1 - p_{\mathrm{accept}}$ (Equation 4). We set $\eta = 2$, which seems to work better in our setting than the default $\eta = 3$. For each position $j \in \{1, ..., T\}$ inside the anchor, Weaver's residual exists only on the candidates $c_j$, so we restrict both distributions to the $K = 512$ candidates and renormalize; the teacher side then costs $K$ dot products and row reads from the frozen vocabulary projection.

Restricting the loss to the top-$K$ tokens is proper as long as only a little of the verifier mass escapes the pool of the top tokens, which DFlash's top-$K$ marginals satisfy in practice[1]. Since greedy verification accepts only on exact argmax agreement, we add a small term to match $\hat{c}_j = \mathrm{argmax}_c\, p_j(c)$, giving the per-position loss:

$$\mathcal{L}_j = L_{\mathrm{LK}}^{\lambda}(p_j, q_j) - \gamma \cdot \log q_j(\hat{c}_j), \quad \gamma = 0.1. \quad (18)$$

During training, we apply the loss only at positions the drafter would have reached under speculative sampling, and mask out positions beyond the first reject. We find that this approach improves performance, which we attribute to its similarity to curriculum learning.

The final loss averages $\mathcal{L}_j$ over only the reached positions of all anchors in a batch.

#### 4.1.4 Dataset

We train on a mixture of instruction-following, chat, and coding data: the Nemotron Post-Training Dataset V2[39], LMSYS-Chat-1M[40], OpenHermes 2.5[41], and CodeAlpaca[42]. We subsample the union to 300k completions and train for a single epoch.

#### 4.1.5 Parameters

Weaver has 56.7M trainable parameters, arranged as a single Transformer layer of dimension 2048 with 16 attention heads and an MLP of width 2048; the can-didate pool size is $K = 512$. We train the large weight matrices with Muon[43] and the remaining parameters with AdamW[44], under a warmup–stable–decay (WSD) schedule with a peak learning rate of $2 \times 10^{-4}$.

[1]Measured on held-out data, the verifier places on average 97.8% of its probability mass inside the top-512 pool, so the out-of-pool mass poses no practical concern in our setting.

## 4.2 Experiments setup

We evaluate the drafter performance on Qwen3.6-27B using an NVIDIA B200 at batch size 1, over workloads that determine user-facing interactivity, in three categories: chat (MTBench[37], ShareChat[45]), math (GSM8K[46], MATH500[47], AIME25[48]), and code (HumanEval[49], MBPP[50], LiveCodeBench[51]).

The DFlash and DDTree baselines use the same publicly released Qwen3.6-27B-DFlash checkpoint that Weaver is trained on. The model card notes that the Qwen3.6 drafter is still under training, so we read the absolute DFlash numbers as a conservative Qwen3.6 baseline.

For a fair comparison across methods, all experiments run in SGLang[12], using our own fork on **GitHub**. DFlash uses its existing SGLang implementation, and we implement DFlash-TfM and the DDTree baseline on top of the same framework. We exclude EAGLE-3 from the comparison because no open-source weights are available for Qwen3.6.

We evaluate proposal quality using tokens/step ($\tau$), the average number of generated tokens per verification cycle, and report end-to-end speedup over the autoregressive baseline in Table 2.

## 4.3 Verification setup

We evaluate each method under different speculative budgets and verification methods, and report the configuration that maximizes the speedup in Table 2.

For the chain proposals, we compare DFlash and DFlash-TfM with budget 16, where DFlash-TfM drafts autoregressively and builds no tree. It is plausible that the fair evaluation protocol is to apply the same sampling policy to the verifier and the drafter for both methods and to verify with speculative sampling. However, because DFlash's argmax proposals are deterministic, naive verification is a natural (and better-performing) fit for them, and we additionally report it in Table 3. In Table 2 we report the best choice for each method: naive verification for DFlash[1] and speculative sampling for DFlash-TfM.

[1]This also follows the default SGLang implementation of DFlash.

For the tree setting, we run DDTree and DFlash-TfM with budgets {32, 64, 128, 256, 512}. On the sampling path, DFlash-TfM verifies the proposal tree with Traversal verification[10], since each draft token is sampled stochastically conditioned on the preceding tokens. For DDTree we keep the verification procedure of the original work[11], so its numbers reflect its intended protocol.

# 5 Experiments

## 5.1 Main results

| | | CHAT | | | | MATH | | | | | | CODE | | | | | | | |
|---|---|---|---|---|---|---|---|---|---|---|---|---|---|---|---|---|---|---|---|
| Path | Method | MTBench | | ShareChat | | GSM8K | | MATH500 | | AIME25 | | HumanEval | | MBPP | | LCB | | Macro Avg. | |
| | | Speedup | τ | Speedup | τ | Speedup | τ | Speedup | τ | Speedup | τ | Speedup | τ | Speedup | τ | Speedup | τ | Speedup | τ |
| **Temperature=0.0** | | | | | | | | | | | | | | | | | | | |
| **No Reasoning** | | | | | | | | | | | | | | | | | | | |
| **Chain** | DFlash (16) | 2.52× | 4.82 | 2.05× | 2.94 | 4.52× | 7.10 | 5.01× | 7.63 | 4.69× | 7.05 | **5.77×** | **11.48** | 4.04× | 7.70 | 3.51× | 5.84 | 4.01× | 6.82 |
| **Chain** | DFlash-TfM (16) | 2.28× | 5.33 | 1.83× | 3.24 | 4.17× | 7.94 | 4.59× | 8.51 | 4.27× | 7.78 | 5.16× | 12.07 | 3.71× | 8.42 | 3.21× | 6.48 | 3.65× | 7.47 |
| **Tree** | DDTree (64) | 2.69× | 6.32 | 2.23× | 4.25 | 4.36× | 9.03 | 4.81× | 9.58 | 4.40× | 8.78 | 5.09× | 12.77 | 3.93× | 9.52 | 3.56× | 7.64 | 3.88× | 8.49 |
| **Tree** | DFlash-TfM (64) | **2.78×** | **7.28** | **2.24×** | **4.76** | **4.65×** | **10.56** | **5.14×** | **11.06** | **4.88×** | **10.48** | 5.10× | 13.61 | **4.09×** | **10.73** | **3.94×** | **9.02** | **4.10×** | **9.69** |
| **Reasoning** | | | | | | | | | | | | | | | | | | | |
| **Chain** | DFlash (16) | 3.06× | 5.01 | 2.44× | 3.70 | 4.37× | 6.83 | 5.09× | 7.35 | 4.77× | 6.85 | 4.43× | 6.36 | 4.27× | 6.20 | 3.72× | 5.38 | 4.02× | 5.96 |
| **Chain** | DFlash-TfM (16) | 2.81× | 5.65 | 2.21× | 4.15 | 4.11× | 7.83 | 4.71× | 8.26 | 4.43× | 7.70 | 4.14× | 7.22 | 3.97× | 7.04 | 3.46× | 6.07 | 3.73× | 6.74 |
| **Tree** | DDTree (64) | 3.17× | 6.63 | 2.61× | 5.16 | 4.33× | 8.76 | 4.76× | 9.13 | 4.49× | 8.55 | 4.34× | 8.27 | 4.21× | 8.06 | 3.69× | 7.10 | 3.95× | 7.71 |
| **Tree** | DFlash-TfM (64) | **3.38×** | **7.91** | **2.71×** | **6.04** | **4.78×** | **10.61** | **5.30×** | **11.05** | **5.12×** | **10.60** | **4.85×** | **10.07** | **4.67×** | **9.86** | **4.13×** | **8.63** | **4.37×** | **9.35** |
| **Temperature=1.0** | | | | | | | | | | | | | | | | | | | |
| **No Reasoning** | | | | | | | | | | | | | | | | | | | |
| **Chain** | DFlash (16) | 2.29× | 4.50 | 1.84× | 2.58 | 4.22× | 6.80 | 4.66× | 7.24 | 4.08× | 6.25 | **5.09×** | **11.07** | 3.59× | 7.32 | 3.03× | 5.19 | 3.60× | 6.37 |
| **Chain** | DFlash-TfM (16) | 2.13× | 4.80 | 1.67× | 2.87 | 3.89× | 7.25 | 4.24× | 7.69 | 3.84× | 6.87 | 4.98× | 11.59 | 3.38× | 7.77 | 2.91× | 5.80 | 3.38× | 6.83 |
| **Tree** | DDTree (64) | 2.47× | 5.96 | 1.96× | 3.73 | 4.03× | 8.65 | 4.45× | 9.13 | 4.03× | 8.17 | 4.90× | 12.45 | 3.57× | 9.04 | 3.17× | 6.88 | 3.57× | 8.00 |
| **Tree** | DFlash-TfM (64) | **2.61×** | **6.59** | **2.07×** | **4.16** | **4.56×** | **9.99** | **4.99×** | **10.49** | **4.58×** | **9.43** | 4.93× | 13.31 | **3.87×** | **10.24** | **3.45×** | **8.03** | **3.88×** | **9.03** |
| **Reasoning** | | | | | | | | | | | | | | | | | | | |
| **Chain** | DFlash (16) | 2.70× | 4.44 | 2.14× | 3.23 | 3.99× | 6.29 | 4.33× | 6.31 | 3.97× | 5.71 | 3.96× | 5.68 | 3.86× | 5.57 | 3.12× | 4.48 | 3.51× | 5.21 |
| **Chain** | DFlash-TfM (16) | 2.53× | 4.98 | 1.96× | 3.55 | 3.79× | 7.06 | 4.23× | 7.32 | 3.83× | 6.53 | 3.83× | 6.54 | 3.67× | 6.34 | 3.05× | 5.26 | 3.36× | 5.95 |
| **Tree** | DDTree (64) | 2.88× | 6.08 | 2.30× | 4.60 | 3.95× | 8.19 | 4.24× | 8.20 | 3.97× | 7.57 | 3.98× | 7.57 | 3.89× | 7.44 | 3.23× | 6.20 | 3.56× | 6.98 |
| **Tree** | DFlash-TfM (64) | **3.16×** | **7.10** | **2.49×** | **5.27** | **4.67×** | **9.86** | **4.90×** | **9.76** | **4.60×** | **9.07** | **4.59×** | **9.07** | **4.49×** | **8.92** | **3.72×** | **7.47** | **4.08×** | **8.31** |

Table 2: Decoding speedup over standard autoregressive decoding and average acceptance length (τ) on Qwen3.6-27B. We follow the DFlash evaluation convention. Parenthesized values indicate the chain block size for DFlash and DFlash-TfM (Chain), and tree budget for DDTree and DFlash-TfM (Tree). Macro Avg. reports the average across datasets. Tree-verified DFlash-TfM outperforms the runner-up (DDTree) by 6% to 15%, depending on the setting.

DFlash-TfM (Tree) is the fastest configuration on every task (Table 2), with an average interactivity of 366.2 tok/s: a 4.08 × speedup over the bfloat16 autoregressive baseline, 16.3% ahead of the tuned DFlash chain. The advantage is largest on MATH500 and smallest on ShareChat, which tracks how predictable the target's outputs are. The gap comes from significantly improved mean acceptance length: Weaver's trees lengthen MAL by 73.6% relative to the chain DFlash baseline and by 22.3% relative to DDTree at the same tree size.

The two baselines isolate the two contributing parts of the method. The com-parison with DDTree isolates the conditional residual: both build trees from the same DFlash marginals, and the 22.3% difference in acceptance length is the part contributed by Weaver's conditioning. The comparison with the chain isolates the tree: with the same drafter, the block-16 chain commits 5.95 tokens per step (301.9 tok/s) against the budget-64 tree's 8.31 (366.2 tok/s), and a longer chain block would not close this gap, since chain acceptance length saturates with draft length, as can be seen from Figure 6.

## 5.2 Kernel Performance

Relative to the per-branch recurrent baseline of Figure 5, the masked solve removes the recomputation of the state along every branch, and the saving grows with the tree size. From table Figure 5 we see that by $T = 128$ the recurrent path is 7.1 × slower than the fused kernel. Inside the kernel, the Gram computation and the solve are the expensive stages, and the solve cost grows super-linearly in $T$ as can be seen from Table 1.

End to end, the GDN verify is only about 12% of the total decode step execution time, 2.5 out of 21 ms. The remainder of the step is the rest of the target forward in attention layers, MLPs, and the non-verify parts of the GDN blocks ($\approx$ 13 ms); and the draft and Weaver preparation (5.1 ms).

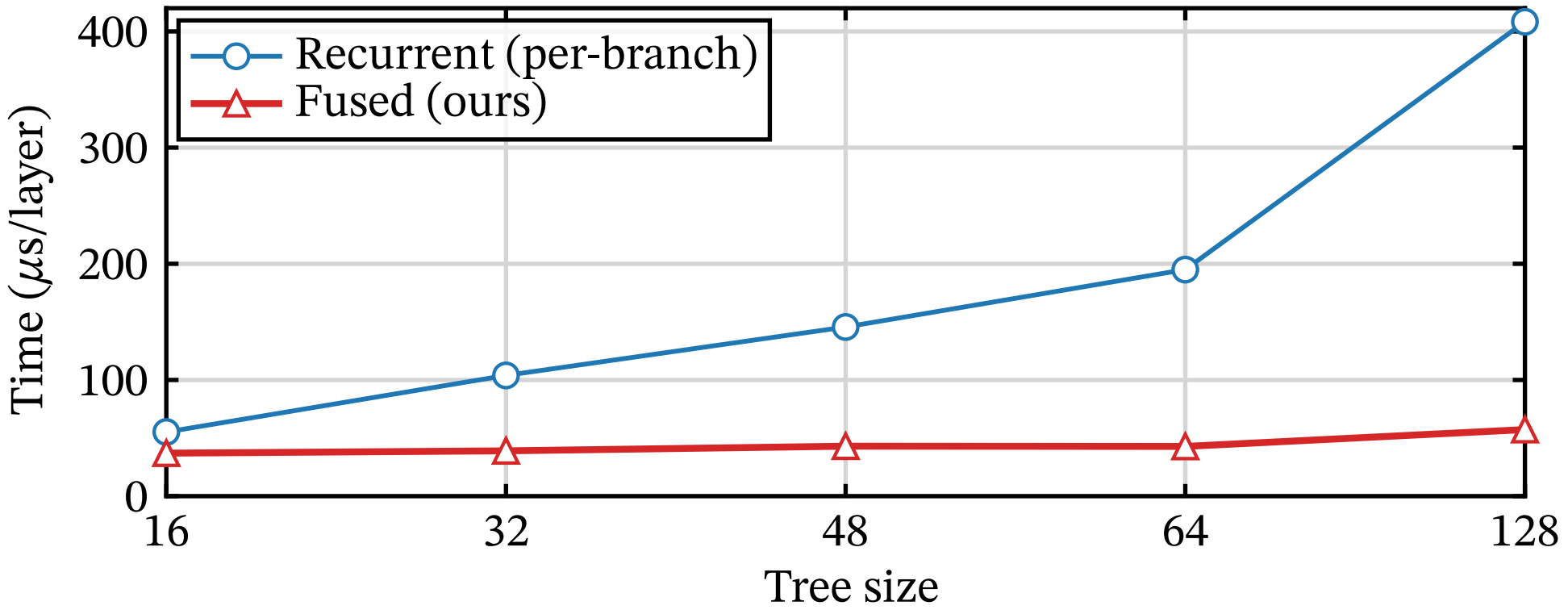

Figure 5: GDN forward pass time on B200, batch size 1

## 5.3 Verification ablation

To identify the coupling scheme that maximizes the performance for DFlash and DFlash-TfM at temperature 1.0, we evaluate naive verification and speculative sampling using budget-16 chain proposals on MTBench with reasoning enabled.

| Method | Proposal | Verification Method | Speedup | $\tau$ |
|---|---|---|---|---|
| DFlash | $\text{argmax}(P_{\text{marg}})$ | Naive Verification | **2.67×** | **4.44** |
| DFlash-TfM | $\text{argmax}(P_{\text{AR}})$ | Naive Verification | 2.42× | 4.90 |
| DFlash | $x \sim P_{\text{marg}}$ | Speculative Sampling | 2.37× | 4.06 |
| DFlash-TfM | $x \sim P_{\text{AR}}$ | Speculative Sampling | **2.48×** | **5.00** |

Table 3: Decoding speedup over standard autoregressive decoding and average acceptance length ($\tau$) on MTBench dataset with Qwen3.6-27B in bfloat16 at temperature 1.0 and reasoning enabled. While DFlash-TfM variant has higher across-the-board acceptance length, in the Naive Verification setting its token generation speed is worse due to the tree construction overhead.

The optimal choice of coupling depends on the drafter type, as can be seen from Table 3. For the marginal drafter DFlash, Naive Verification gives higher tokens/step than speculative sampling. In contrast, the autoregressive drafter DFlash-TfM shows the opposite result. The gain in tokens/step is reflected in the speedup. We therefore conclude that Naive Verification is the best scheme for DFlash and use it throughout our experiments.

To see where the difference comes from, we also plot the mean accepted length as a function of a draft length in a Figure 6.

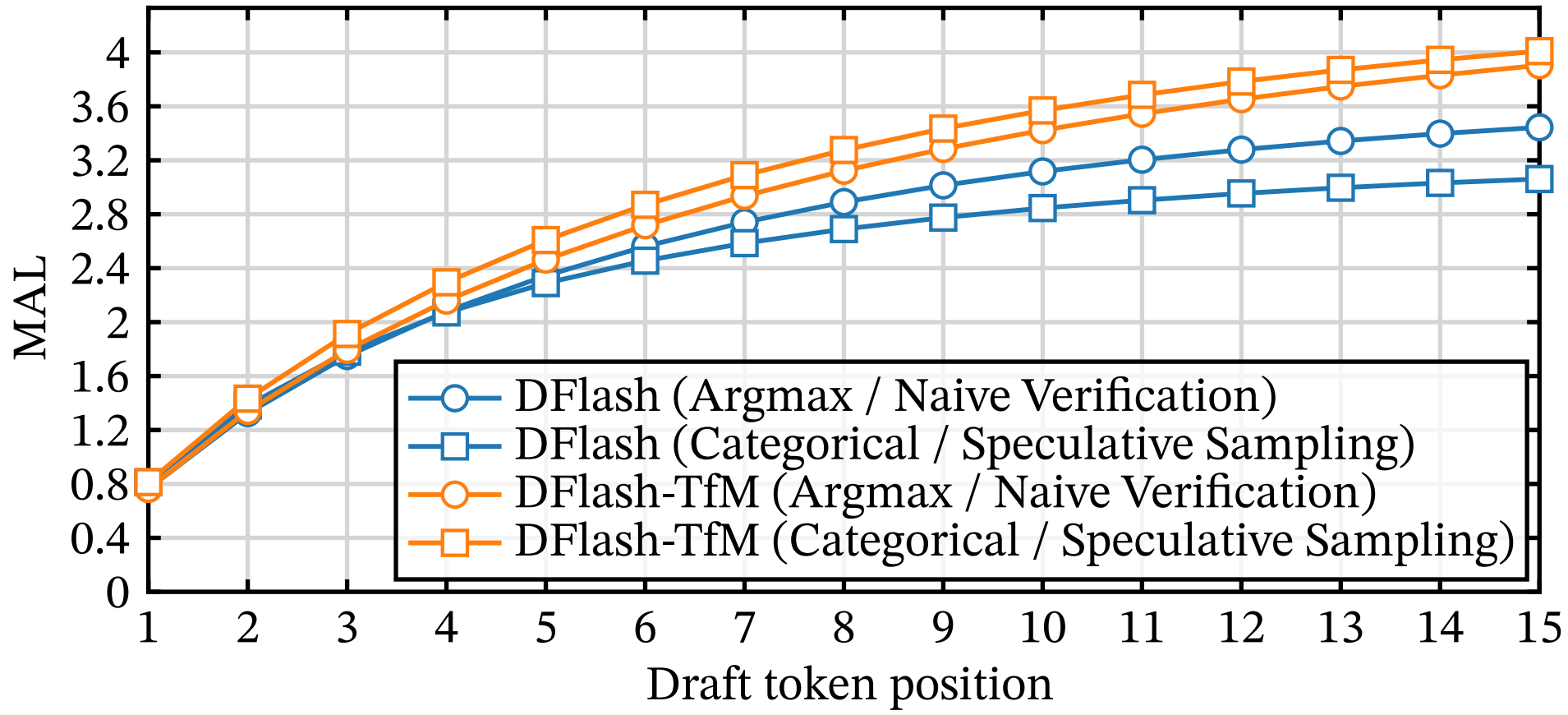

Figure 6: Mean accepted length versus draft length for chain proposals on MTBench (Qwen3.6-27B, bfloat16, temperature 1.0, reasoning on, budget 16). DFlash argmax+naive overtakes categorical+speculative as draft length grows; DFlash-TfM shows no such crossover.

This follows the behavior in Figure 4. Speculative sampling (i.e. the proposal and verifier are sampled by the same policy) is better near the start of the draft, but the curves cross around draft lengths 2–4, After that point, argmax proposals under Naive Verification provide longer acceptance length.

We hypothesize that at short draft lengths sampled proposals still approximate the verifier distribution well and the prefixes diverge little. The effect grows with draft length, as each token conditions on a longer generated prefix and staying on a high-probability path matters more.

DFlash-TfM does not show this crossover because Weaver predicts the next token from the previously sampled draft tokens.

# 6 Conclusion and future work

We showed that the acceptance ceiling of factorized drafters follows from their independence assumption rather than from drafter capacity, and that a small autoregressive adapter is sufficient to lift it. Weaver conditions DFlash's marginals on the realized draft tokens using 56.7M parameters and no multiplication by the full-vocabulary projection, and the resulting proposal trees exceed the acceptance rates available to any marginal-only drafter at long depths. On the systems side, the proposed kernel reduces tree verification for delta-rule layers to a masked triangular solve: the committed state is never speculatively written, so no rollback is required, and the construction applies to any target built on non-diagonal linear attention. Together these contributions yield a $4.08\times$ speedup over autoregressive decoding and 16.3% over the previous state-of-the-art, a tuned DFlash baseline.

Several future directions are feasible. Tree construction maximizes draft probability, which is a proxy for acceptance; constructing and pruning the tree against an estimate of the acceptance probability itself would spend the same budget on a higher quality tree. The draft depth is limited by DFlash's block of 16 positions; longer marginal windows, or re-anchoring the drafter on its own proposals mid-draft, would let the tree grow past this horizon. Weaver is trained against a frozen drafter; training the two jointly could shape the marginal support around the conditional corrections it feeds. Finally, the verify kernels already operate on batched state while the serving path handles one request at a time; extending the scheduler to concurrent requests would carry the method from the interactivity regime into throughput serving.

# Appendix

## 7 Pseudocode for tree construction

Algorithm 1: Adaptive tree construction. The marginal prior and the Weaver context cache are computed once per speculation step; the tree is then grown best-first under budget $B$ and expansion width $w$, with the per-node fan-out fixed at $c = 8$. Functions and containers are capitalized; scalars and record fields are lowercase.

```
function BuildTree(B, w, c, K, h_t, r_t)
    input: budget B, expansion width w, fan-out c, candidate pool size K, verifier state h_t, bonus token r_t
    (h'_{t+1}, …, h'_{t+L}) ← MarginalDrafter(h_t)
    for l = 1 to L do
        (cands_{t+l}, priors_{t+l}) ← TopK(W_lm h'_{t+l}, K)      ▷ Top-K tokens and logits
    context ← ContextTokens(h_t, h'_{t+1}, …, h'_{t+L})
    rootcache ← Project(context)                                ▷ KV cache shared by all steps
    MaxHeap.push((token : r_t, prob : 1.0, cache : rootcache))
    Tree ← ∅
    for i = 1 to ⌈B/w⌉ do
        Batch ← MaxHeap.PopTop(w)                               ▷ w most probable frontier nodes
        Tree.Add(Batch)
        (hidden, newcaches) ← WeaverStep(Batch.tokens, Batch.caches)   ▷ Batched step
        for each node ∈ Batch do
            residual ← (W_q RMSNorm(hidden_node))^T W_lm [cands_node]  ▷ Depth-l candidates
            probs ← softmax(priors_node + residual)             ▷ Corrected distribution
            for each child ∈ Top(c, probs) do
                MaxHeap.push((token : child, prob : probs[child] · node.prob, cache : newcaches_node))
    return Tree
```

## 8 Budget Sweep

This appendix reports the full budget sweep not shown in Table 2. We plot interactivity (tokens/sec/sequence) against speculative budget under each sampling and reasoning setting. Within each dataset group, bars are ordered by increasing budget: DDTree {32, 64, 128, 256, 512} and DFlash-TfM {16, 32, 64, 128, 256, 512}; budget 16 corresponds to the chain proposal. The AR baseline and DFlash (budget 16) are single configurations.

#### 8.0.1 Sweep (temperature 1.0, reasoning on)

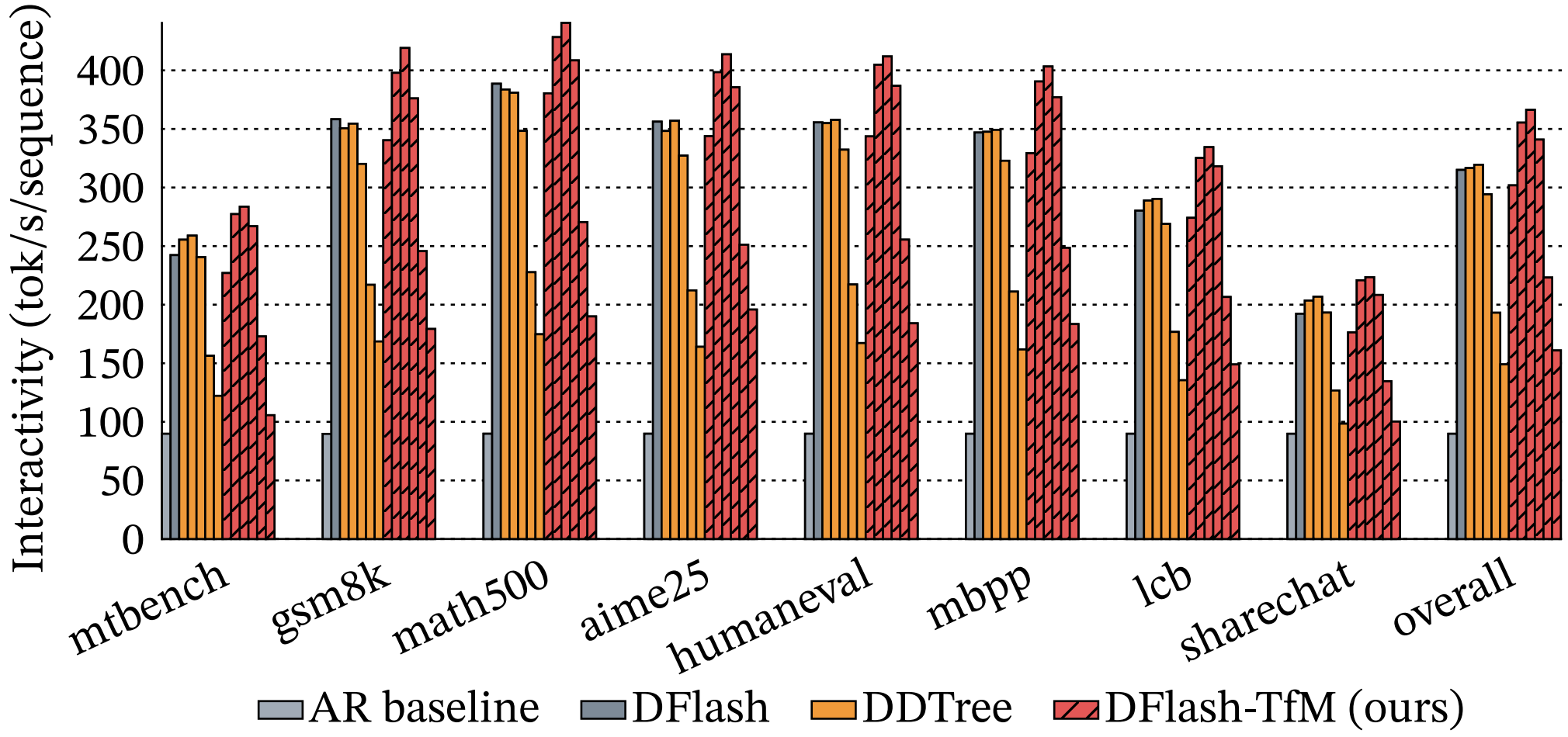

Figure 7: Interactivity versus speculative budget with temperature 1.0 and reasoning on.

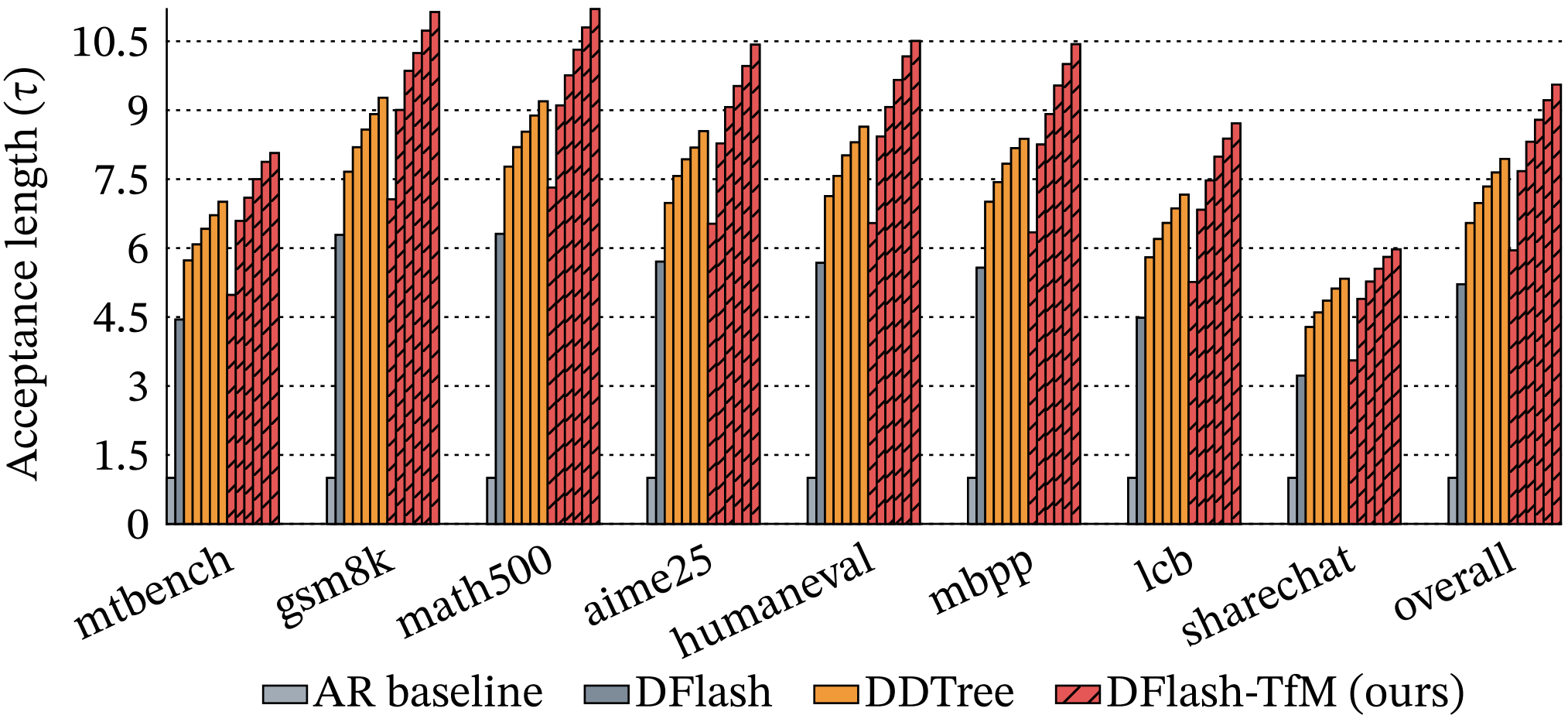

Figure 8: Acceptance length (τ) versus speculative budget temperature 1.0 and reasoning on.

### 8.0.2 Sweep (temperature 1.0, reasoning off)

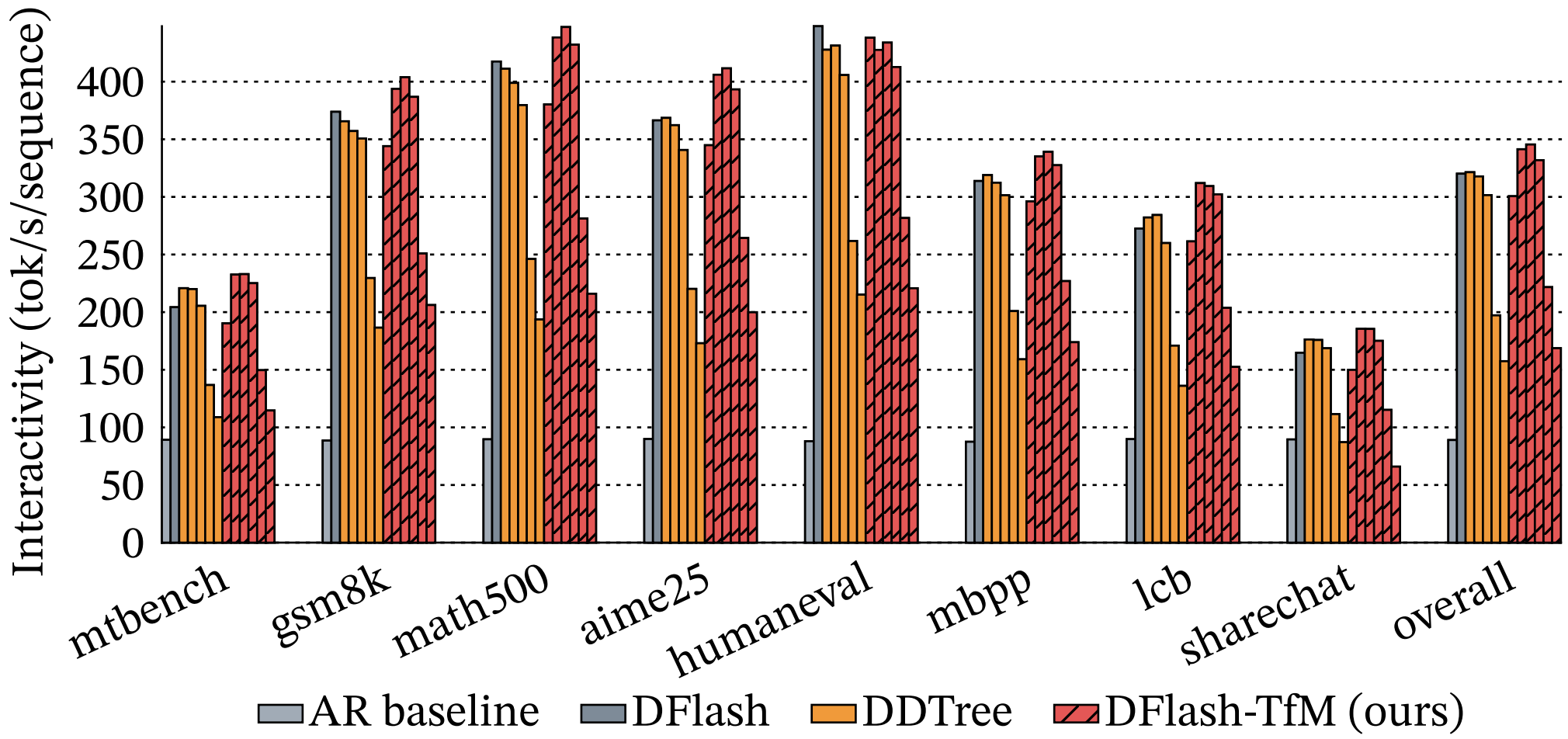

Figure 9: Interactivity versus speculative budget with temperature 1.0 and reasoning off.

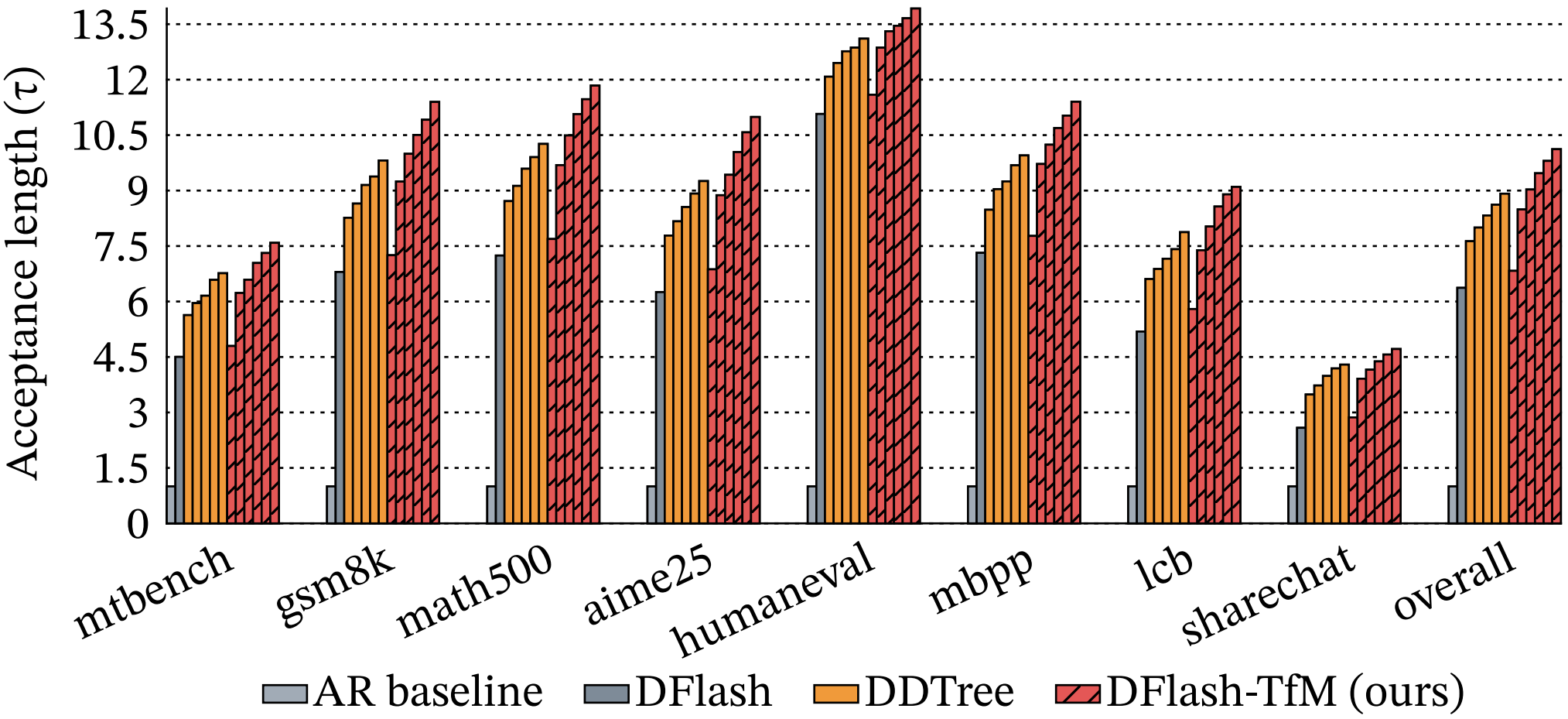

Figure 10: Acceptance length (τ) versus speculative budget temperature 1.0 and reasoning off.

### 8.0.3 Sweep (greedy decoding, reasoning on)

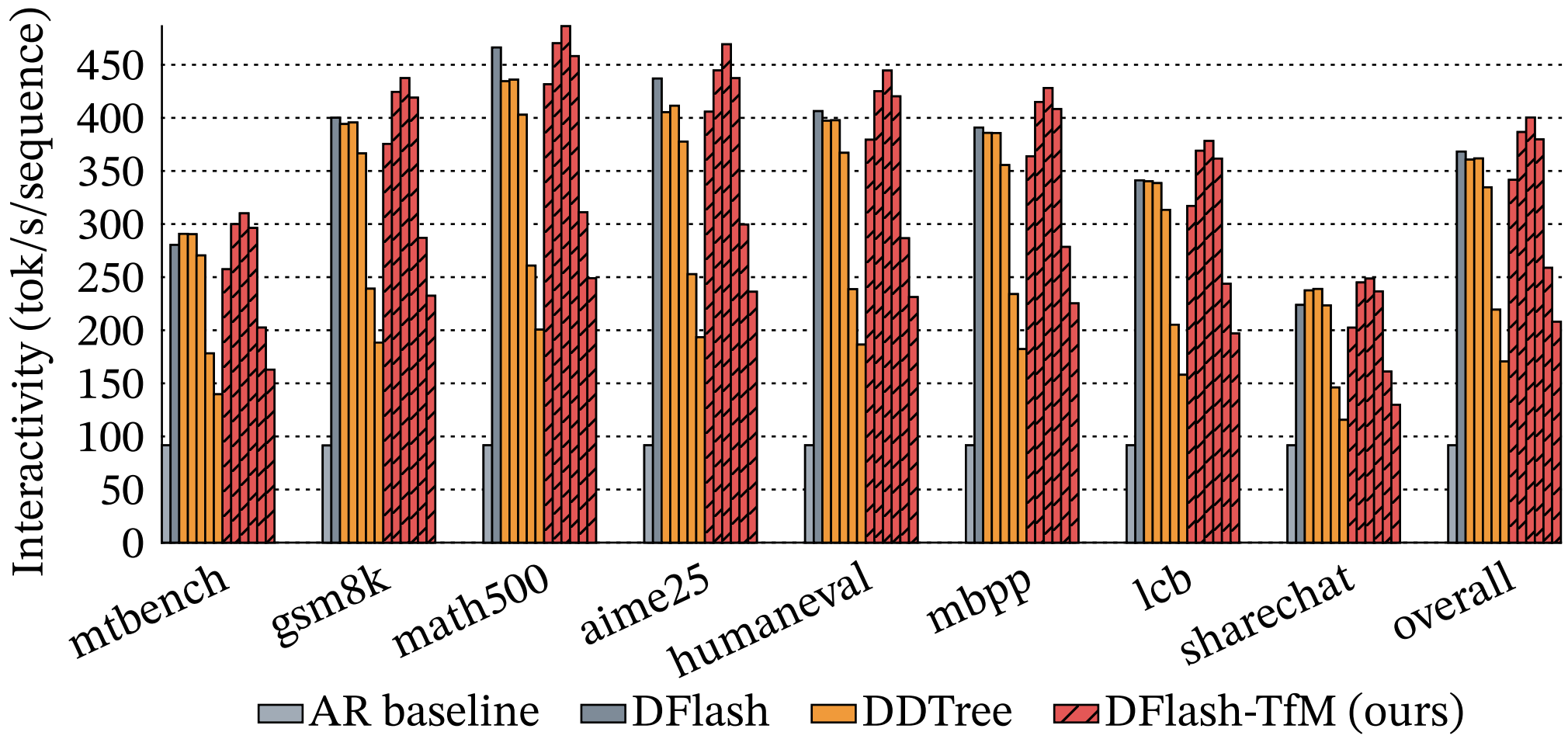

Figure 11: Interactivity versus speculative budget with greedy decoding and reasoning on.

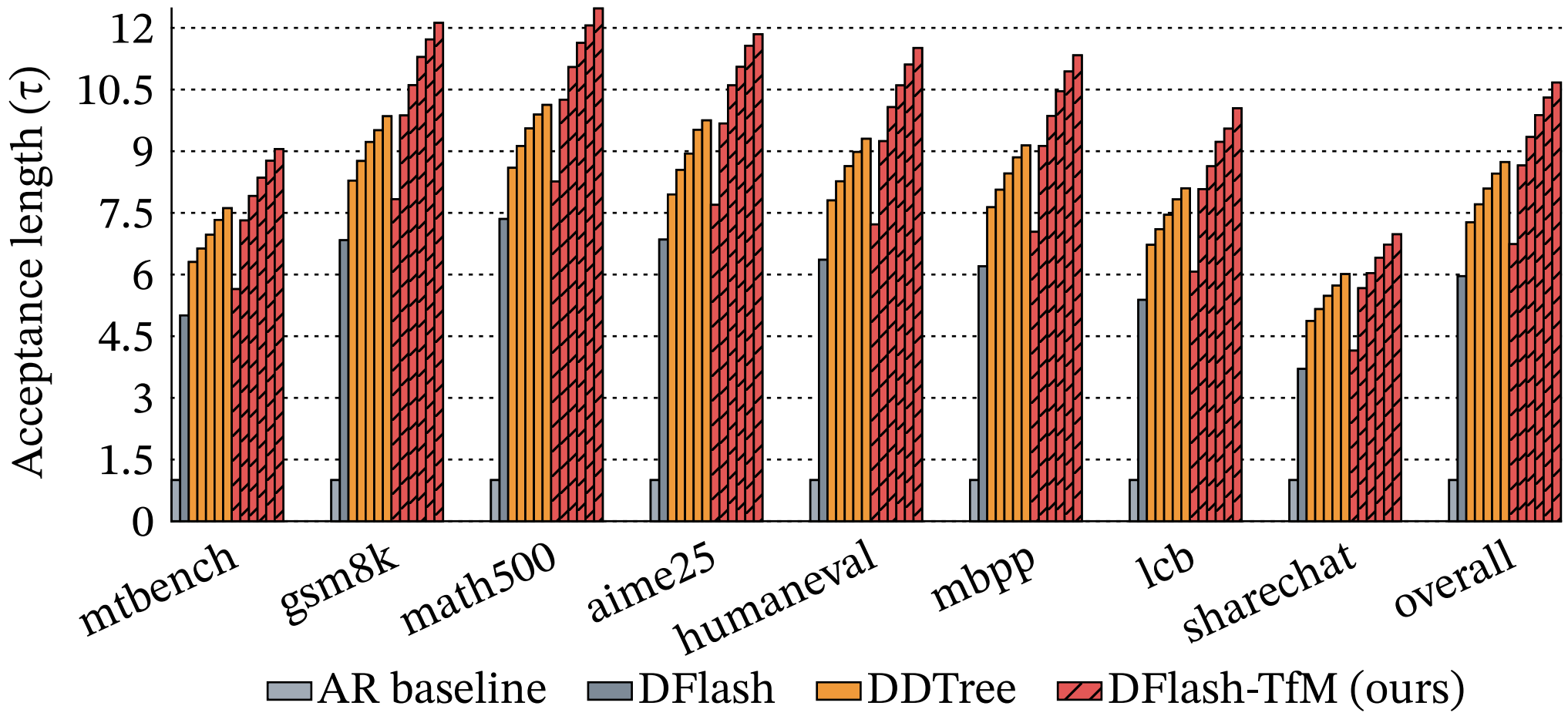

Figure 12: Acceptance length ($\tau$) versus speculative budget greedy decoding and reasoning on.

### 8.0.4 Sweep (greedy decoding, reasoning off)

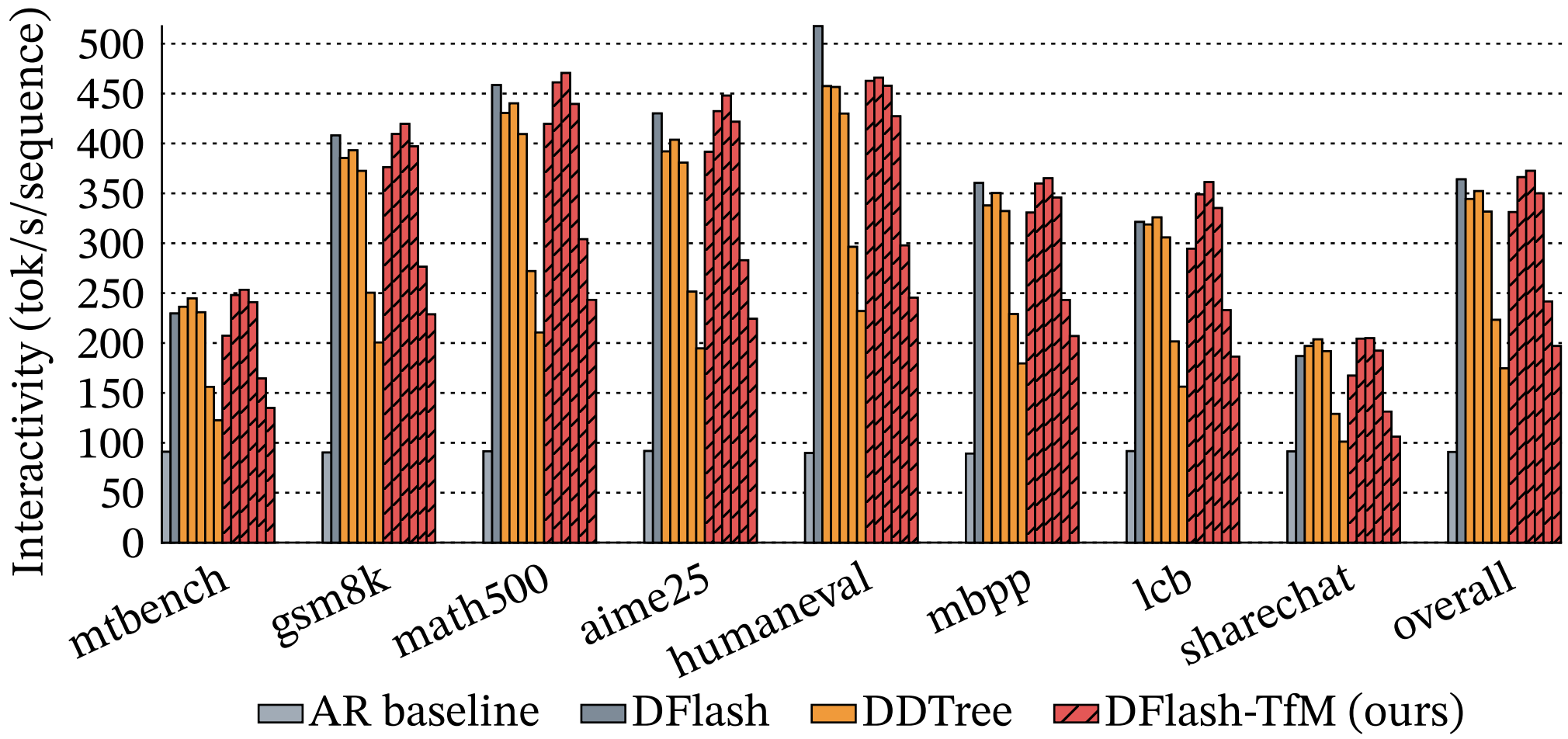

Figure 13: Interactivity versus speculative budget with greedy decoding and reasoning off.

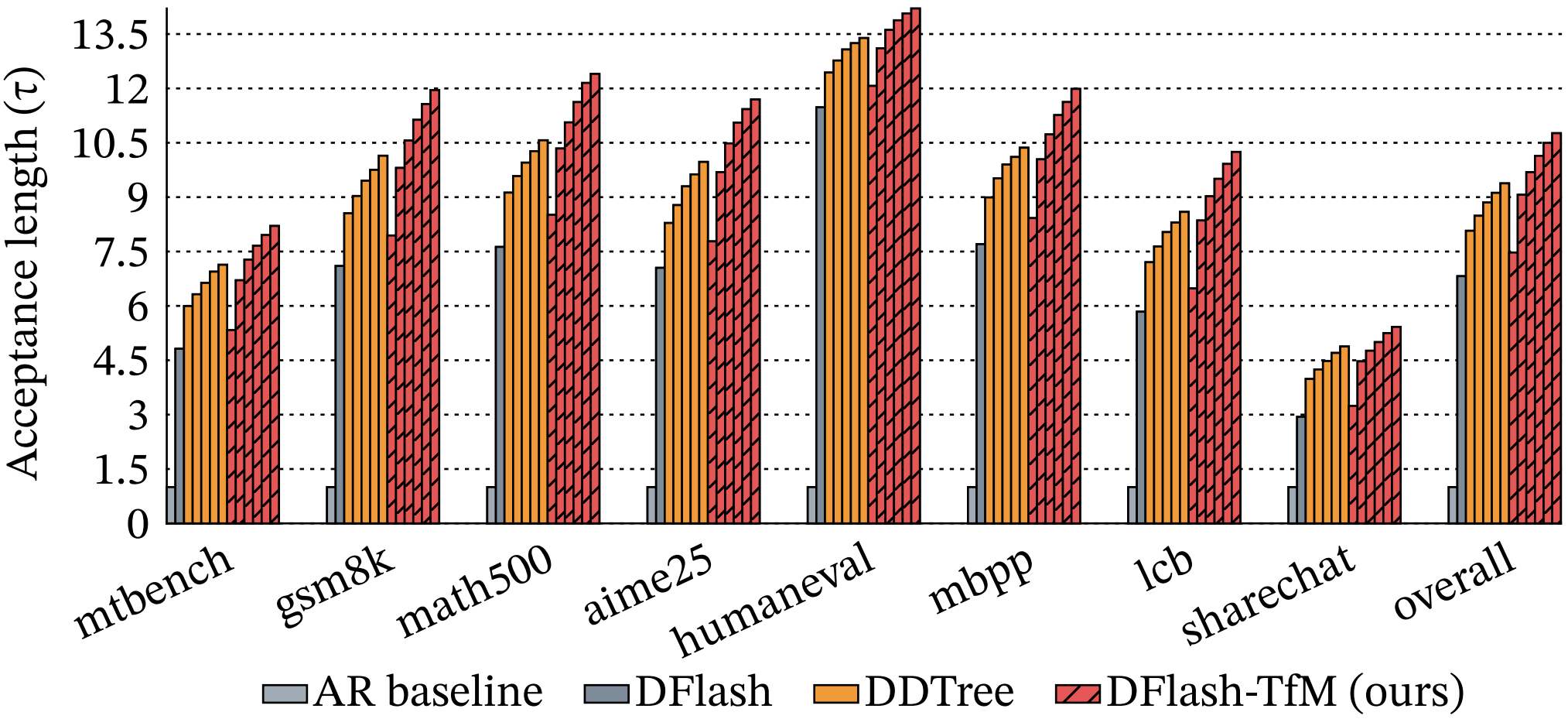

Figure 14: Acceptance length ($\tau$) versus speculative budget greedy decoding and reasoning off.

# 9 Acceptance rate of DFlash variants

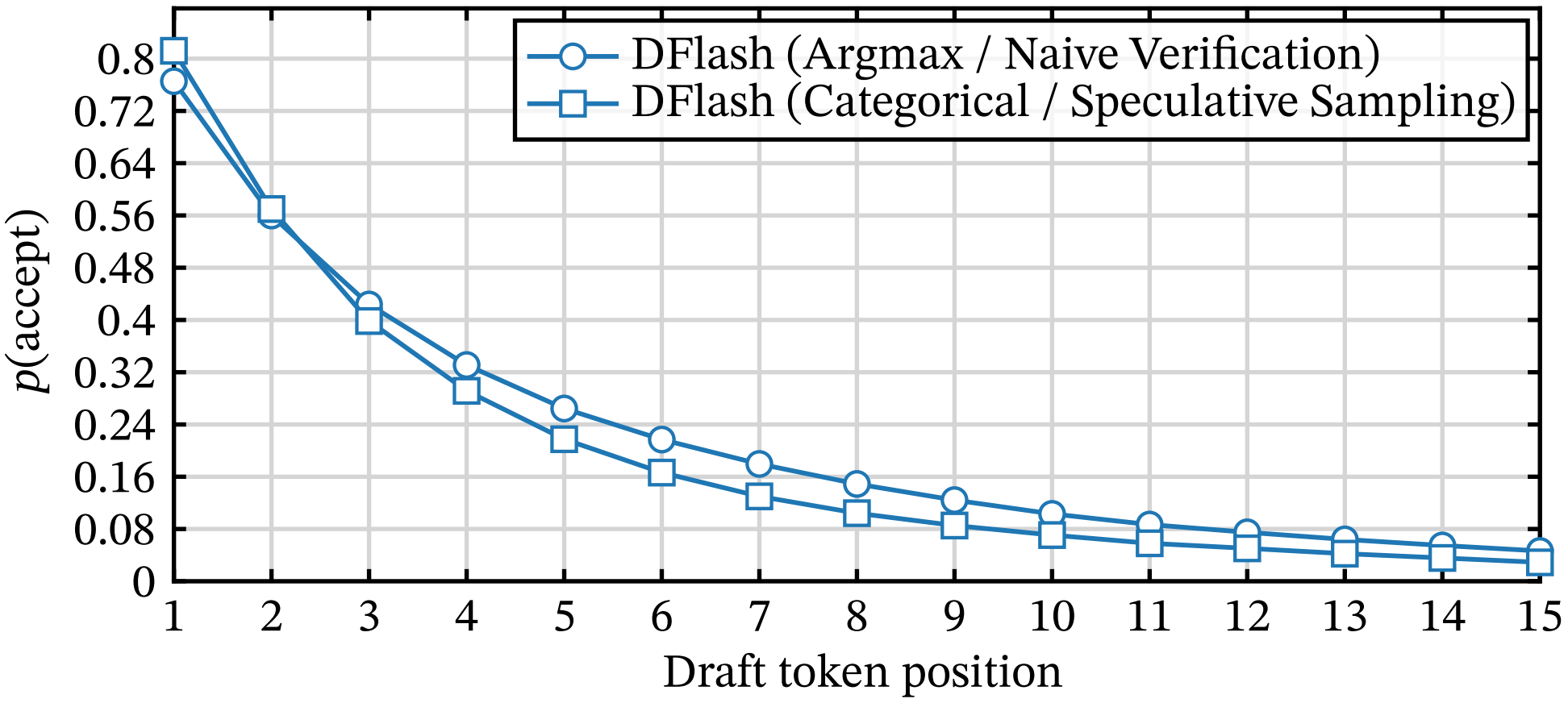

Figure 15: DFlash: acceptance by draft token position on MTBench with temperature 1.0

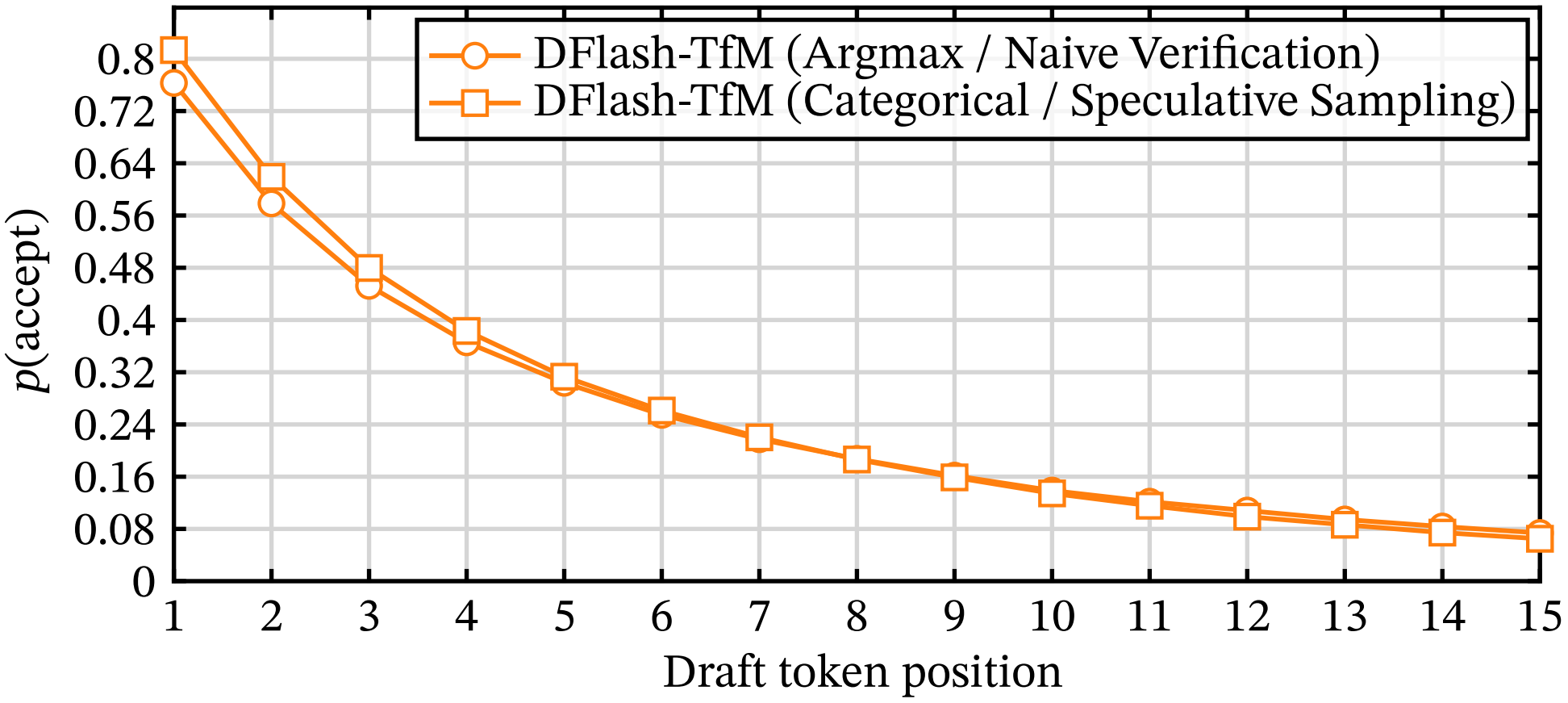

Figure 16: DFlash-TfM: acceptance by draft token position on MTBench with temperature 1.0